\documentclass[sigconf,nonacm]{acmart}
\AtBeginDocument{%
  }

\setcopyright{rightsretained}
\copyrightyear{2026}
\acmYear{2026}
\acmDOI{}
\acmConference[COLIEE 2026]{COLIEE 2026 workshop}{June 12,
  2026}{Singapore}
\acmBooktitle{Proceedings of COLIEE 2026 workshop, June 12, 2026, Singapore}
\acmISBN{}
\acmDOI{}

\usepackage{subcaption}
\usepackage{float}
\usepackage{bbm}
\usepackage[normalem]{ulem}
\useunder{\uline}{\ul}{}

\usepackage[most]{tcolorbox}
\begin{document}

\title{NOWJ@COLIEE 2026: Adaptive Pipelines for \\Legal Retrieval and Reasoning}


\author{Thuong-Hieu Ngo}
\affiliation{%
  \institution{VNU University of Engineering and
Technology}
  \city{Hanoi}
  \country{Vietnam}}
\email{25025063@vnu.edu.vn}

\author{Hoang-Trung Nguyen}
\affiliation{%
  \institution{VNU University of Engineering and
Technology}
  \city{Hanoi}
  \country{Vietnam}}
\email{25025010@vnu.edu.vn}

\author{Huu-Dong Nguyen}
\affiliation{%
  \institution{VNU University of Engineering and
Technology}
  \city{Hanoi}
  \country{Vietnam}}
\email{21020760@vnu.edu.vn}

\author{Xuan-Bach Le}
\affiliation{%
  \institution{VNU University of Engineering and
Technology}
  \city{Hanoi}
  \country{Vietnam}}
\email{22024506@vnu.edu.vn}

\author{Le-Dung Nguyen}
\affiliation{%
  \institution{VNU University of Engineering and
Technology}
  \city{Hanoi}
  \country{Vietnam}}
\email{24022633@vnu.edu.vn}

\author{Quang-Thanh Tran}
\affiliation{%
  \institution{VNU University of Engineering and
Technology}
  \city{Hanoi}
  \country{Vietnam}}
\email{24021625@vnu.edu.vn}

\author{Ha-Thanh Nguyen}
\affiliation{%
  \institution{Center for Juris-Informatics, ROIS-DS \\ College of Engineering \& Computer Science, VinUniversity}
  \city{Hanoi}
  \country{Vietnam}}
\email{thanh.nh6@vinuni.edu.vn}

\author{Thi-Hai-Yen Vuong}
\affiliation{%
  \institution{VNU University of Engineering and
Technology}
  \city{Hanoi}
  \country{Vietnam}}
\email{yenvth@vnu.edu.vn}

\renewcommand{\shortauthors}{Thuong-Hieu Ngo et al.}

\begin{abstract}
This paper presents the methodologies and results of the NOWJ team's participation across all five tasks of the COLIEE 2026 competition. For Task~1 (Legal Case Retrieval), we propose a four-stage pipeline comprising candidate filtering, dense retrieval with complementary embedding models, cross-encoder reranking via fine-tuned generative rerankers and MLP-based pairwise classification, and adaptive per-query cutoff prediction. For Task~2 (Legal Case Entailment), we combine BM25 filtering, T5-based reranking, and LLM-based entailment verification with consensus ensemble. For Task~3 (Statute Law Retrieval and Entailment), we adopt a retrieval-augmented generation framework with dense retrieval, attention-based reranking, and few-shot prompted LLM reasoning. For Task~4 (Legal Textual Entailment), we introduce a dynamic routing pipeline that classifies query difficulty and dispatches cases to either a balanced few-shot solver or a structured zero-shot chain-of-thought solver. For the Pilot Task (Legal Judgment Prediction), we combine hierarchical transformers with CRF layers, argument relation mining, and probabilistic argumentation graph reasoning.
\end{abstract}




\maketitle

\section{Introduction}
Legal text processing is a specialized field that requires expertise in both law and information science. The increasing adoption of artificial intelligence (AI) and large language models (LLMs) in judicial processes has driven growing interest in automating core legal tasks such as case retrieval, statutory reasoning, and judgment prediction. The Competition on Legal Information Extraction and Entailment (COLIEE)~\cite{rabelo2024coliee} is an annual shared task organized to advance research in this area, covering challenges ranging from document retrieval and textual entailment to judgment prediction. COLIEE 2026, the thirteenth edition of the competition, comprises five tasks that span case law, statute law, and tort law.

Task~1 (Legal Case Retrieval) requires retrieving noticed cases from a corpus of case law documents that support a given query case whose citations have been redacted. Task~2 (Legal Case Entailment) requires identifying which paragraph in a noticed case entails the decision fragment of the query case. Task~3 (Statute Law Retrieval and Entailment) requires systems to jointly return an entailment decision and the supporting articles from a collection of Japanese Civil Code articles for a given legal query. Task~4 (Legal Textual Entailment) requires determining whether a given set of gold relevant articles entails or contradicts a legal query. The Pilot Task (LJPJT26) requires predicting whether a tort is affirmed and extracting the accepted arguments from both plaintiff and defendant claims.

This paper presents the NOWJ team's participation across all five tasks. For Task~1, we propose a four-stage pipeline of candidate filtering, dense retrieval, cross-encoder reranking with two complementary approaches, and adaptive per-query cutoff prediction. For Task~2, we employ BM25 filtering, T5-based reranking, and LLM-based entailment verification with consensus ensemble. For Task~3, we adopt a retrieval-augmented generation framework combining dense retrieval, attention-based reranking, and few-shot prompted LLM reasoning. For Task~4, we introduce a dynamic routing pipeline that classifies query difficulty and dispatches cases to specialized solvers. For the Pilot Task, we combine hierarchical transformers with CRF layers, argument relation mining, and probabilistic argumentation graph reasoning. Our system achieves competitive performance across all tasks, with particularly strong results on Task~1 and Task~3, demonstrating the effectiveness of multi-stage pipelines that integrate dense retrieval with task-specific reranking and reasoning strategies. The remainder of this paper is organized by task, with each section detailing the task formulation, our proposed methodology, experimental setup, and results.

\section{Task 1: Legal Case Retrieval}
\subsection{Task Overview}
The rapid growth of legal databases has made manual identification of relevant precedents increasingly impractical. COLIEE Task~1 addresses this by formulating legal case retrieval as follows: given a query case $q$ with all citations redacted, the objective is to retrieve all noticed cases $\{d_1, d_2, \ldots, d_n\}$ from a corpus of case law documents, each containing the case background, facts, reasoning, and decision. The noticed cases are those originally cited in support of the decision in $q$. The task presents several challenges: case documents typically range from 4,000 to 60,000 words, exceeding the context window of many transformer-based models; documents lack standardized structure; and the corpus contains mixed-language content (English and French).

In COLIEE 2025, the top-performing team, JNLP~\cite{jnlp2025}, achieved an F1 of 0.3353 by extending the UMNLP pairwise ranking framework with BM25 and SAILER-based features in a feed-forward classifier, with BM25 pre-filtering achieving 76--85\% recall on the top 100--200 candidates. The runner-up, UQLegalAI~\cite{uqlegalai2025}, employed CaseLink, a graph neural network modeling case-level connectivity through contrastive learning, attaining an F1 of 0.2962. A consistent trend across participants was the adoption of multi-stage pipelines combining sparse retrieval with neural re-ranking and LLM-based summarization for document compression.

\subsection{Methodology}

\begin{figure}[!t]
    \centering
    \includegraphics[width=1\linewidth]{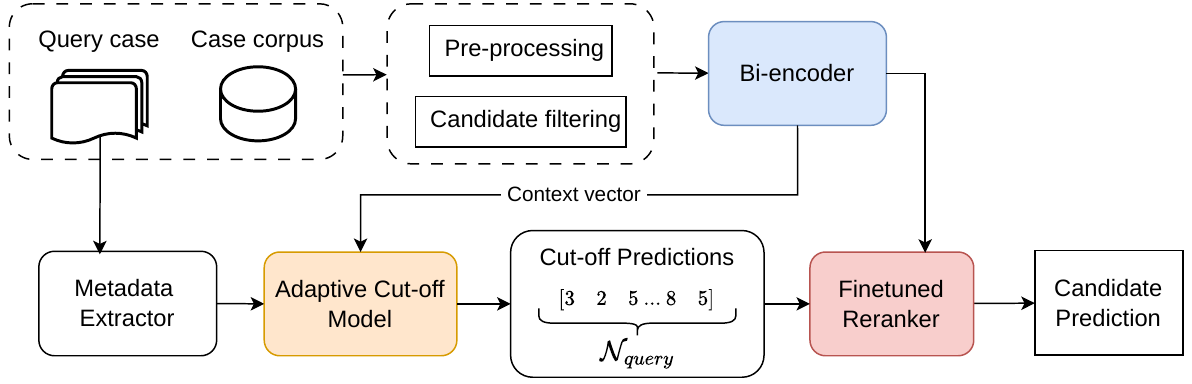}
    \caption{Four-stage pipeline including preprocessing and candidate filtering, dense retrieval, reranking, and adaptive per-query cutoff for legal case retrieval.}
    \label{fig:task1_framework}
    \Description{Task 1 framework}
\end{figure}

To address the key challenges of Legal Case Retrieval---namely the excessive document length and complex logical structure of case texts---we propose a four-stage framework. The pipeline first applies year-based filtering to exclude temporally invalid candidates, followed by dense retrieval for initial ranking and reranking for refined relevance scoring. The overall architecture is illustrated in Figure~\ref{fig:task1_framework}.

\paragraph{\textbf{Stage 1: Preprocessing and Candidate Filtering}}
Since the Federal Court of Canada corpus contains bilingual documents, French-language content is first removed using an automatic language detector\footnote{\url{https://huggingface.co/papluca/xlm-roberta-base-language-detection}}. We then apply two filtering heuristics: (i) excluding candidates whose trial date is later than the query's, as valid noticed cases must temporally precede the query, and (ii) removing cases that also appear as queries, as these are rarely cited as noticed cases. These steps prevent temporally invalid citations and substantially narrow the candidate space.

\paragraph{\textbf{Stage 2: Dense Retrieval}}
The dense retrieval stage identifies the most promising candidates from the full corpus, balancing computational efficiency with high recall. We adopt two complementary pre-trained embedding models: Octen-Embedding-8B and E5-Mistral-7B-Instruct. Each document is independently encoded into dense vectors by both models. Following an instruction-aware encoding paradigm, queries are augmented with task-specific instructions to convey the retrieval objective, while documents are encoded in a neutral format.
 
\begin{tcolorbox}[
  colback=gray!10,
  colframe=black!50,
  boxrule=0.5pt,
  arc=3pt,
  left=2pt, right=2pt, top=2pt, bottom=2pt
]
\ttfamily
\textbf{Query:} \texttt{Instruct: Given a legal document, retrieve relevant legal documents\textbackslash nQuery: \{query text\}} \\[3pt]
\textbf{Document:} \texttt{passage: \{document text\}}
\end{tcolorbox}

\noindent Since case documents frequently exceed the embedding model's context window, both queries and documents are split into chunks and independently encoded. To compute document-level relevance, we adopt a MaxSim scoring function inspired by ColBERT~\cite{khattab2020colbert}:
\begin{equation}
    s(q, d) = \sum_{i=1}^{M} \max_{j \in [N]} \mathbf{q}_i \cdot \mathbf{d}_j
\end{equation}
where $\{\mathbf{q}_i\}_{i=1}^{M}$ and $\{\mathbf{d}_j\}_{j=1}^{N}$ denote the query and document chunk embeddings, respectively. This formulation allows each query chunk to match its most relevant document segment, effectively capturing partial relevance across long documents. All embeddings are $\ell_2$-normalized, making the dot product equivalent to cosine similarity. The top-$k$ results are forwarded to the reranking stage.

\paragraph{\textbf{Stage 3: Reranking}}

Following the candidate retrieval stage, we introduce a reranking module to refine the ranking of candidate cases. This stage enhances retrieval performance by incorporating more expressive relevance modeling beyond the initial similarity scores.
We propose two independent reranking approaches, each capturing different aspects of semantic and structural relevance between legal cases.

\subparagraph{\textbf{Approach 1: Weighted Ensemble.}}
Dense retrieval encodes queries and candidates independently, limiting fine-grained semantic matching. The reranking stage addresses this by jointly encoding each query-candidate pair, enabling token-level interaction for more accurate relevance estimation.

We employ Qwen3-Reranker~\cite{qwen3reranker}, a generative reranker that produces relevance scores from the logits at the final token position:

\begin{equation}
    s(q, d) = \frac{\exp(l_{\text{yes}})}{\exp(l_{\text{yes}}) + \exp(l_{\text{no}})}
\end{equation}
where $l_{\text{yes}}$ and $l_{\text{no}}$ are the logits of the ``yes'' and ``no'' tokens, casting relevance assessment as a binary judgment over the generated token distribution.
 
Both Qwen3-Reranker-4B and 8B variants are fine-tuned on COLIEE 2026 training data using ms-swift~\cite{msswift} with pointwise binary cross-entropy loss:

\begin{equation}
    \mathcal{L} = -\left[y \log s(q, d) + (1 - y) \log (1 - s(q, d))\right]
\end{equation}
where $y \in \{0, 1\}$  is the ground-truth relevance label. Positive samples are drawn from gold noticed cases. To construct informative negatives, we apply a denoising strategy that filters out candidates whose retrieval scores fall too close to those of positive samples, ensuring that the training signal is not diluted by ambiguous near-positive examples. All training and inference follow the Qwen3-Reranker prompt template.

\begin{tcolorbox}[
  colback=gray!10,
  colframe=black!50,
  boxrule=0.5pt,
  arc=3pt,
  left=2pt, right=2pt, top=2pt, bottom=2pt
]
\ttfamily
\textbf{System:} Judge whether the Document meets the requirements based on the Query and the Instruct provided. Note that the answer can only be "yes" or "no". \\
\textbf{User:} <Instruct>: Given a legal document, retrieve relevant legal documents \textbackslash n <Query>: \{query text\} \textbackslash n <Document>: \{candidate document text\} \\
\textbf{Assistant:} \{yes / no\}
\end{tcolorbox}

The final relevance score for each candidate integrates signals from all preceding stages---the dense retrieval score and both cross-encoder reranking scores---via a weighted combination:
\begin{equation}
    s_{\text{fused}}(q, d) = \alpha \cdot \text{sim}(q, d) + \beta \cdot s_{\text{4B}}(q, d) + \gamma \cdot s_{\text{8B}}(q, d)
\end{equation}
where $\text{sim}(q, d)$ is the cosine similarity from Stage~2, $s_{\text{4B}}$ and $s_{\text{8B}}$ are the reranking scores from Qwen3-Reranker-4B and 8B respectively, and $\alpha, \beta, \gamma$ are hyperparameters tuned on the development set ($\alpha=0.5$, $\beta=1.0$, $\gamma=1.5$). This multi-signal fusion allows the system to leverage the broad recall of dense retrieval alongside the fine-grained relevance judgments of the cross-encoders.

\subparagraph{\textbf{Approach 2: MLP-based Reranking.}}

To refine the initial retrieval results, we adopt a pairwise classification framework inspired by UMNLP~\cite{umnlp2024}. Given a query–candidate pair, the model estimates a relevance probability using a multi-layer perceptron (MLP).

Candidate Construction.
For each query, we build a candidate pool from the top-200 candidates retrieved in Stage 2 to reduce noise while maintaining high recall. Ground-truth relevant cases are treated as positive instances, while the remaining candidates serve as hard negatives. These top-ranked non-relevant cases are semantically closer to the query than random negatives, encouraging the model to learn finer-grained distinctions in legal reasoning.

Feature Construction.
Each query–candidate pair is represented using proposition-level features following UMNLP. We further incorporate:
\begin{itemize}
    \item \textbf{Semantic features}: Maximum cosine similarity from dense embeddings.
    \item \textbf{Lexical features}: BM25 scores and normalized variants.
    \item \textbf{Rank features}: Rank positions and reciprocal ranks from retrieval models.
    \item \textbf{Metadata features}: Legal domain categories and case types.
\end{itemize}

These features allow the model to jointly capture semantic similarity, lexical overlap, and structural signals from the retrieval stage.

Model Architecture and Training.
The reranker is implemented as a feed-forward neural network that takes the concatenated feature vector and outputs a relevance probability $p(q,d) \in [0,1]$. The model is trained using binary cross-entropy loss:
\begin{equation}
\mathcal{L} = -\left[y \log p(q,d) + (1-y)\log(1-p(q,d))\right],
\end{equation}

We train directly on the original data distribution without aggressive resampling, allowing the model to better reflect real-world retrieval conditions.

\paragraph{\textbf{Stage 4: Adaptive Per-Query Cutoff}}
After reranking, the system must determine how many candidates to return for each query. This is non-trivial because the number of gold noticed cases varies significantly across queries---some have a single relevant precedent while others have several. Applying a fixed threshold would inevitably sacrifice either precision or recall depending on the query. To address this, we train an XGBoost and MLP ensemble to predict the expected number of relevant cases $|\mathcal{D}_q|$ for each query. The predictor takes as input document metadata features (e.g., document length, year) and dense embeddings from Octen-Embedding-0.6B. The predicted count is used to adaptively truncate the ranked list, enabling the system to dynamically adjust its output size per query and better balance precision and recall.

\subsection{Experimental Setup}
Building upon the proposed methodology, our experimental setup instantiates the four-stage pipeline with specific pre-trained models and training configurations. The first two stages---preprocessing with candidate filtering and dense retrieval---are shared across all runs. For the dense retrieval stage, we employ two complementary embedding models: Octen-Embedding-8B\footnote{\url{https://huggingface.co/Octen/Octen-Embedding-8B}} and E5-Mistral-7B-Instruct\footnote{\url{https://huggingface.co/intfloat/e5-mistral-7b-instruct}}, both used without task-specific fine-tuning. The three runs differ in the reranking approach applied from Stage~3 onward.
 
We submitted three official runs for Task~1:
\begin{itemize}
    \item \textbf{Run 1 (submission\_1):} Reranking with fine-tuned Qwen3-Reranker-4B\footnote{\url{https://huggingface.co/Qwen/Qwen3-Reranker-4B}} and Qwen3-Reranker-8B\footnote{\url{https://huggingface.co/Qwen/Qwen3-Reranker-8B}} via weighted score fusion, followed by adaptive per-query cutoff (XGBoost/MLP). Both models are fine-tuned using ms-swift~\cite{msswift} with LoRA~\cite{hu2021lora} (rank 16, $\alpha{=}32$, dropout 0.05) on all linear layers, lr $1{\times}10^{-6}$, batch size 1, max length 16,660/8,400 tokens (4B/8B), 3 epochs with early stopping.
    
    \item \textbf{Run 2 (submission\_2):} Stage-2 candidates are reranked using the MLP-based approach (Approach~2), which combines proposition-level, semantic, lexical, rank, and metadata features. Propositions are extracted with flan-t5-xl\footnote{\url{https://huggingface.co/google/flan-t5-xl}} and semantic embeddings are obtained via multilingual-e5-large-instruct\footnote{\url{https://huggingface.co/intfloat/multilingual-e5-large-instruct}}, with the MLP predicting relevance scores for final reranking.
    
    \item \textbf{Run 3 (submission\_3):} Combines predictions from Run 1 and Run 2 via score-level fusion. The scores from both runs are first normalized and then summed to produce final rankings, giving higher priority to cases consistently ranked highly by both methods.
\end{itemize}

\subsection{Results and Discussion}
Table~\ref{tab:recall_comparison} presents the recall performance of two dense retrieval models, Octen and E5, across multiple cutoff levels ranging from top-1 to top-200. Overall, both models demonstrate strong retrieval capability, with recall consistently improving as the cutoff increases.

\begin{table}[htbp]
\centering
\caption{Recall at top-k cutoffs for Octen-Embedding-8B and E5-Mistral-7B-Instruct on the COLIEE 2026 Task 1 test set.}
\label{tab:recall_comparison}
\begin{tabular}{c c c}
\hline
Top-$k$ & Octen-Embedding-8B & E5-Mistral-7B-Instruct \\
\hline
R@1   & 0.1086 & 0.1103 \\
R@3   & 0.2463 & 0.2440 \\
R@5   & 0.3269 & 0.3309 \\
R@10  & 0.4440 & 0.4474 \\
R@20  & 0.5703 & 0.5680 \\
R@50  & 0.7063 & 0.7079 \\
R@100 & 0.7663 & 0.7863 \\
R@200 & 0.8274 & 0.8463 \\
\hline
\end{tabular}
\end{table}

Table~\ref{tab:task1_results} shows that our proposed four-stage pipeline achieves competitive performance, with \texttt{submission\_2}, \texttt{submission\_3}, and \texttt{submission\_1} ranked 1st, 2nd, and 6th overall, respectively. The reranking stage improves precision by better distinguishing relevant cases from hard negatives. Among reranking strategies, the MLP feature-based reranker outperformed fine-tuned generative rerankers, as it leverages hand-crafted domain-specific features --- such as cited article overlap, charge alignment, and structural case similarities --- that are particularly discriminative in legal retrieval. In contrast, fine-tuned generative rerankers, while powerful in general settings, are optimized as domain-agnostic models intended to generalize across diverse data types and tasks, which may limit their ability to exploit the fine-grained, dataset-specific patterns present in legal corpora. Furthermore, the adaptive per-query cutoff enhances performance by dynamically balancing precision and recall. These findings suggest that combining domain-tailored reranking with an adaptive cutoff is a robust approach for legal case retrieval.

\begin{table}[htbp]
\centering
\caption{Official results of the top runs on COLIEE 2026 Task~1.}
\label{tab:task1_results}
\begin{tabular}{c l l c c c}
\hline
Rank & Team & Run & F1 & Precision & Recall \\
\hline
1 & \underline{\textbf{NOWJ}} & \underline{\textbf{submission\_2}} & \underline{\textbf{0.4220}} & \underline{{0.4235}} & \underline{{0.4206}} \\
2 & \underline{NOWJ} & \underline{submission\_3} & \underline{0.4200} & \underline{0.4140} & \underline{\textbf{0.4263}} \\
3 & JNLP & random\_forest & 0.4126 & \textbf{0.4341} & 0.3931 \\
4 & JNLP & ensemble & 0.4040 & 0.4174 & 0.3914 \\
5 & JNLP & xgboost & 0.4000 & 0.4059 & 0.3943 \\
6 & \underline{NOWJ} & \underline{submission\_1} & \underline{0.3880} & \underline{0.3772} & \underline{0.3994} \\
7 & SIL & submission\_sil2 & 0.3871 & 0.4186 & 0.3600 \\
8 & SIL & submission\_sil1 & 0.3810 & 0.3856 & 0.3766 \\
\hline
\end{tabular}
\end{table}

\section{Task 2: Legal Case Entailment}
\subsection{Task Overview}

Given a decision $d$ and a relevant case $R=\{p_1,p_2,...,p_n\}$, Task 2 focuses on identifying the specific paragraph $p\in R$ that entails $d$. Unlike Task 1, which operates at the document level, this task demands fine-grained comprehension, as multiple paragraphs may discuss related legal concepts without directly supporting the final decision.

\subsection{Methodology}

To address this challenge, our system employs a multi-stage pipeline integrating sparse information retrieval, neural reranking, and LLM-based refinement. The pipeline consists of three core stages:

\textbf{BM25 Initial Filtering:}
We utilize BM25, a sparse retrieval method based on term-frequency inverse-document-frequency (TF-IDF), to retrieve the top-$k$ candidate paragraphs from $R$ using $d$ as the query. 

\textbf{Neural Reranking:}
The candidates are reranked using a T5-based cross-encoder. This model computes a relevance score by capturing deeper semantic interactions between $d$ and $p$:
$$\text{score}(d, p) = \text{CrossEncoder}(d, p)$$
This step promotes contextually aligned paragraphs, filtering out lexical noise from the BM25 stage.

\textbf{LLM Refinement and Ensemble Fusion:} 
To make the final assessment, we deploy state-of-the-art LLMs to evaluate the highest-ranked outputs. Each LLM generates a binary prediction (entails/does not entail) based on specialized prompts. To ensure high precision, we apply a consensus-based ensemble fusion:
$$\text{final}(p) = \text{LLM}_1(p) \land \text{LLM}_2(p)$$
where $\text{final}(p) = 1$ indicates that the paragraph entails the decision. This mechanism mitigates individual model biases and hallucinations.

\subsection{Experimental Setup}

The proposed pipeline is implemented using specific pre-trained models. BM25 is used for initial filtering, followed by the {MonoT5-3B model}\footnote{\url{https://huggingface.co/castorini/monot5-3b-msmarco-10k}} for cross-encoder reranking. For the final LLM refinement, we experiment with {Qwen3-235B-A22B}\footnote{\url{https://huggingface.co/Qwen/Qwen3-235B-A22B}} and {QwQ-32B}\footnote{\url{https://huggingface.co/Qwen/QwQ-32B}}.

To evaluate the effectiveness of individual LLMs and our proposed ensemble strategy, we submitted three official runs:

\begin{itemize}
    \item \textbf{Run 1 (NOWJ001):} A single-LLM pipeline using Qwen3-235B-A22B for the final verification step.
    \item \textbf{Run 2 (NOWJ002):} A similar single-LLM setup, but utilizing QwQ-32B to evaluate its standalone performance.
    \item \textbf{Run 3 (NOWJ003):} The ensemble fusion configuration. A paragraph $p$ is classified as entailment ($\text{final}(p) = 1$) if and only if both Qwen3-235B-A22B and QwQ-32B output positive predictions. Disagreements are discarded to prioritize precision.
\end{itemize}

\subsection{Results and Discussion}

As shown in Table \ref{tab:task2leaderboard}, our ensemble approach (NOWJ/nowj003) achieved an F1 score of 0.4429, ranking fourth overall. A notable characteristic across all our submissions is the strong emphasis on precision. Run 1 (NOWJ/nowj001) recorded the highest precision among all participating teams at 0.7604, significantly outperforming the top-ranked IAI team (0.4501). Runs 2 and 3 also maintained precision above 0.70. This indicates that our two-stage reranking and LLM-based verification pipeline is highly effective at filtering out false positives. When the system predicts entailment, its confidence is consistently high.

\begin{table}[htbp]
\caption{Leaderboard of the Case Textual Entailment task.}
\label{tab:task2leaderboard}
\begin{tabular}{llll}
\hline
\textbf{Team} & \multicolumn{1}{c}{\textbf{F1}} & \multicolumn{1}{c}{\textbf{Precision}} & \multicolumn{1}{c}{\textbf{Recall}} \\ \hline
\textbf{IAI} & \textbf{0.4899} & 0.4501 & \textbf{0.5374} \\
AIIRLab & 0.4706 & 0.5120 & 0.4354 \\
JNLP & 0.4507 & 0.4174 & 0.4898 \\
{\ul NOWJ/nowj003} & {\ul 0.4429} & {\ul 0.7037} & {\ul 0.3231} \\
UA & 0.4254 & 0.4711 & 0.3878 \\
{\ul NOWJ/nowj002} & {\ul 0.4029} & {\ul 0.7034} & {\ul 0.2823} \\
JUNLLP & 0.3942 & 0.6721 & 0.2789 \\
{\ul NOWJ/nowj001} & {\ul 0.3744} & {\ul \textbf{0.7604}} & {\ul 0.2483} \\ \hline
\end{tabular}
\end{table}

However, this high precision comes at the expense of recall. Our recall scores ranged from 0.2483 to 0.3231, noticeably lower than those of the leading teams. This suggests that the LLMs applied overly strict criteria, resulting in a high number of false negatives. In addition, the initial BM25 retrieval stage may have filtered out lexically different but semantically relevant passages before they could be evaluated by the LLMs.

Comparing our internal runs further confirms the effectiveness of the ensemble approach. Using only the large Qwen3-235B model (Run 1) maximized precision but led to the lowest recall. The smaller QwQ-32B model (Run 2) provided a better balance, achieving a higher F1 score. Ultimately, the ensemble configuration (Run 3) proved to be the most effective setup. It stabilized the predictions, achieving the highest overall F1 and recall scores without significantly reducing precision.

For future improvements, the primary focus will be on increasing recall. This can be addressed by passing a larger candidate pool $k$ to the cross-encoder or by relaxing the LLM prompts to capture broader entailment patterns while preserving our precision advantage.

\section{Task 3: Statute Law Retrieval}
\subsection{Task Overview}

Task~3 aims to retrieve a relevant subset of statutory articles for a legal query $q$ from a fixed collection of Japanese civil law articles $A=\{a_1, a_2, \dots, a_m\}$. In the end-to-end setting of COLIEE~2026, the system also returns a binary answer $\hat{y}\in\{Y,N\}$ supported by the predicted article set $\hat{A}_q \subseteq A$. The training data consist of triplets $(q, A_q, y)$, where $A_q$ is the gold set of relevant articles and $y$ is the Yes/No label.

Most previous systems for statute retrieval in COLIEE follow a multi-stage pipeline. Previous systems typically begin with a retrieval step using lexical methods or compact encoder models, then apply reranking or LLM-based filtering to improve the quality of the selected articles \cite{goebel2025overview}.
Besides text matching, recent studies show that using the structural relations between legal articles can further improve performance. Vuong et al. \cite{vuong2025uncovering} show that modeling law articles as a reference network helps capture both local and long-range dependencies. This approach improves retrieval accuracy and reduces noise. These findings suggest that strong end-to-end performance depends not only on finding relevant articles, but also on filtering out irrelevant ones before making the final prediction.

\subsection{Methodology}

The system uses a three-stage retrieval-and-reasoning pipeline for Task~3. The goal is to automatically predict an entailment decision $\hat{y} \in \{Y,N\}$ together with a subset of supporting statutory articles $\hat{A}_q \subseteq A$ for a given query $q$.

\textbf{Dense Retrieval.}
The first stage aims to identify a subset of relevant articles $\hat{A}_q$ from the full collection $A$. A pre-trained embedding model encodes both the input query and all legal articles into a shared vector space. Similarity between the query and each article is computed to rank the candidates. The top-ranked articles are selected as the retrieved set $\hat{A}_q$, which serves as the supporting evidence for the final decision.

\textbf{Prompt Construction.}
To support the reasoning process, the system constructs a few-shot prompt using training data. For a given query $q$, similar queries in the training set are identified based on embedding similarity. Their corresponding triplets $(q, A_q, y)$ are used as examples, where both the gold supporting articles $A_q$ and the entailment label $y$ are included. These examples are formatted into a prompt to guide the model in producing both a decision and the final article set.

\textbf{Entailment Prediction.}
In the final stage, the model receives the query $q$, the retrieved article set $\hat{A}_q$, and the constructed prompt. Based on this input, it predicts the entailment label $\hat{y} \in \{Y,N\}$ and may refine the selected articles before producing the final output. The output thus aligns with the task requirement: generating both a binary decision and a subset of articles that support it.

\subsection{Experimental Setup}

Building on the proposed methodology, we use a unified retrieval-and-reasoning pipeline for the Statute Law Retrieval task. Given a query $q$ and a statutory corpus $A$, the system aims to produce both an entailment decision $\hat{y} \in \{Y, N\}$ and a subset of supporting articles $\hat{A}_q \subseteq A$.

In the retrieval stage, \texttt{Qwen/Qwen3-Embedding-8B} is used to encode both queries and statutory articles into a shared vector space. Dense retrieval is applied using cosine similarity, and the top 20 articles are selected as the initial supporting set $\hat{A}_q$.

For Run~1, this set is further refined using QRHead \cite{zhang2025query} reranking. QRHead reranks candidates based on attention signals from the model, prioritizing articles more relevant to the query. Articles with higher attention scores are placed earlier, as they are more relevant to the query. This step refines the quality of the final supporting set $\hat{A}_q$.

In the reasoning stage, large language models take the query $q$ and the retrieved article set $\hat{A}_q$ as input to produce the final decision $\hat{y}$. Prompt design is used to guide the reasoning process, and some runs include few-shot examples in the form of $(q, A_q, y)$.

We submitted three official runs:

\begin{itemize}

    \item \textbf{Run 1 (NOWJ\_run1):} After the initial retrieval stage, QRHead is applied to re-rank the top 20 articles. The re-ranked article set $\hat{A}_q$ is then provided to \texttt{Qwen/Qwen3-235B-A22B} with a few-shot prompt for final answer prediction.

    \item \textbf{Run 2 (NOWJ\_run2):} The top-20 retrieved articles are used without reranking. The reasoning stage uses a single model, \texttt{deepseek-ai/DeepSeek-R1-0528}, with a few-shot prompt (Prompt~1) to produce $\hat{y}$.

    \item \textbf{Run 3 (NOWJ\_run3):} The same retrieval setup as Run~2 is used. In the reasoning stage, three different prompts (Prompt~1, Prompt~2, and Prompt~3) are applied using the same model. Each prompt produces a prediction, and the final decision $\hat{y}$ is chosen by majority voting. 

\end{itemize}

\subsection{Results and Discussion}

As shown in Table~\ref{tab:task3_results}, our system performs well across all three runs regarding entailment accuracy, with clear differences between the retrieval strategies. Run~1 (NOWJ\_run1) achieves the highest accuracy of 0.9512 with 78 correct predictions, ranking first overall. This demonstrates that adding QRHead reranking after the initial dense retrieval helps the LLM better focus on the most relevant articles. Runs~2 and~3, which omit the reranking step, achieve a lower accuracy of 0.8902. Furthermore, the identical results between Run~2 (single prompt) and Run~3 (majority voting) indicate that ensemble prompting offers negligible benefits when the underlying retrieved context remains unimproved.

\begin{table}[htbp]
\centering
\caption{Leaderboard of the Statute Law Retrieval task.}
\label{tab:task3_results}
\begin{tabular}{l c c c c}
\hline
Run & Accuracy & Correct & F2 & Recall \\
\hline
\underline{NOWJ/NOWJ\_run1} & \textbf{0.9512} & \textbf{78} & 0.2224 & \textbf{1.0000} \\
JNLP1 & 0.9024 & 74 & \textbf{0.7796} & \textbf{1.0000} \\
\underline{NOWJ/NOWJ\_run2} & 0.8902 & 73 & 0.2224 & \textbf{1.0000} \\
\underline{NOWJ/NOWJ\_run3} & 0.8902 & 73 & 0.2224 & \textbf{1.0000} \\
codyrag & 0.8537 & 70 & 0.1848 & \textbf{1.0000} \\
\hline
\end{tabular}
\end{table}

While our system achieves perfect recall (1.0000) across all runs, the notably low F2 score (0.2224) highlights a clear limitation in the article selection stage. LLMs like Qwen and DeepSeek demonstrate strong deductive reasoning for the binary entailment task (achieving up to 0.9512 accuracy), yet they tend to adopt a conservative approach in evidence extraction. Rather than isolating the exact one or two necessary articles, the LLMs retain additional articles that provide background context but are not strictly legally required to formulate the $\hat{A}_q$ subset.

Compared to the JNLP1 baseline, which maintains perfect recall alongside a high F2 score (0.7796), it is evident that our current pipeline lacks a dedicated filtering mechanism. Relying solely on prompt-based extraction causes the LLM to prioritize the final Yes/No prediction over extraction precision. This suggests that future improvements should decouple the reasoning stage into two distinct phases: a strict statutory filtering phase to refine $\hat{A}_q$, followed by a separate entailment prediction phase.

\section{Task 4: Legal Textual Entailment}

\subsection{Task Overview}
Task~4 of the COLIEE focuses on Legal Textual Entailment. The primary objective is to construct a question-answering framework tailored for the legal domain. Formally, given a legal query $q$ alongside a set of relevant statutory articles $A = \{a_1, a_2, \dots, a_k\}$ (where $k \ge 1$), and a training dataset structured as triplets $(q, A, L)$ with the target label $L \in \{Y, N\}$, the system must output a binary decision indicating whether $A$ legally entails $q$. This task serves as a critical benchmark for investigating the capabilities of computational models in performing complex legal reasoning and precise statutory interpretation.

Reflecting on the previous iteration, the prevailing trend among top participants was the strategic adoption of Large Language Models (LLMs) to handle complex legal reasoning. The winning team, KIS~\cite{kis2025}, achieved top performance primarily by leveraging effective \textit{few-shot prompting} strategies with a Japanese LLM. Meanwhile, the runner-up, CAPTAIN~\cite{captain2025}, enhanced their system by utilizing a \textit{cause-effect} structural analysis tailored for civil law to generate structured examples, which were then integrated using data combination techniques. Furthermore, to prevent test-set data leakage, the competition strictly mandates that all submissions employ open-source models released prior to a specific cutoff date.

\subsection{Methodology}
In this task, we designed a dynamic, routing-based pipeline. Rather than applying a uniform approach, our method processes each legal query through a sequence of logical steps, adapting its reasoning strategy based on the inherent difficulty of the task.

\paragraph{Step 1: Difficulty Classifier}
When a query and its relevant statutory articles enter our system, the first step is to evaluate their structural and logical complexity. We achieve this through a hybrid classifier. Initially, we apply rule-based heuristics: if the articles contain specific keywords indicating mutatis mutandis application, exceptions, or if the query requires synthesizing information from multiple articles, we immediately flag it as a ``Hard'' case. If these rules are not triggered, a lightweight LLM prompt is used as a secondary filter to scan for convoluted conditions or double negations, ultimately classifying the query as either ``Easy'' or ``Hard''. The prompt used for this step is as follows:


\begin{tcolorbox}[
  colback=gray!10,
  colframe=black!50,
  boxrule=0.5pt,
  arc=3pt,
  left=2pt, right=2pt, top=2pt, bottom=2pt
]
\ttfamily
You are a legal task difficulty assessor.\\[0.5em]
Analyze the following legal clause and question, and determine "Hard" if it is logically complex, or "Easy" if it is a simple matching.\\[0.5em]
Hard: Negation of negation, subject mismatch, complex conditions. If unsure, choose "Hard".\\
Easy: Simple question directly from the legal clause.\\
Answer "difficulty: Hard" if the query is Hard, and "difficulty: Easy" if it is Easy.\\[0.5em]
Relevant articles: \{articles\_text\}\\[0.5em]
Question: \{query\_text\}
\end{tcolorbox}

\paragraph{Step 2: Dynamic Routing and Solving}
Based on the difficulty assessment, the query is routed to one of two specialized solvers:

\begin{itemize}
    \item \textit{For ``Easy'' Cases (Dynamic Balanced Few-Shot Solver):} Drawing inspiration from KIS\cite{kis2025}, we optimize for pattern matching here. We embed the input using the \textit{intfloat/multilingual-e5-large\footnote{\url{https://huggingface.co/intfloat/multilingual-e5-large}}} model and retrieve the most semantically similar examples from the training set. To prevent label bias, we enforce strict class balancing, retrieving exactly $k=6$ ``Yes'' and $k=6$ ``No'' examples. These are injected into the prompt, allowing the LLM to make a straightforward binary decision using in-context learning.
    


    \item \textit{For ``Hard'' Cases (Zero-Shot Chain-of-Thought Solver):} Requiring rigorous legal deduction, we frame the LLM as a logical analyst required to articulate its steps within \texttt{<reasoning>} tags. The model first executes \textbf{Rule Extraction and Reference Resolution} to decompose statutes, followed by \textbf{Fact Subsumption} to map assertions against these rules. Additionally, strict guardrails prevent common fallacies (e.g., hallucinations, misuse of contrapositives, or misinterpreting provisos). Upon detecting any structural mismatch or logical leap, the model rejects the premise when structural mismatches are detected.
    

    \begin{figure}[t]
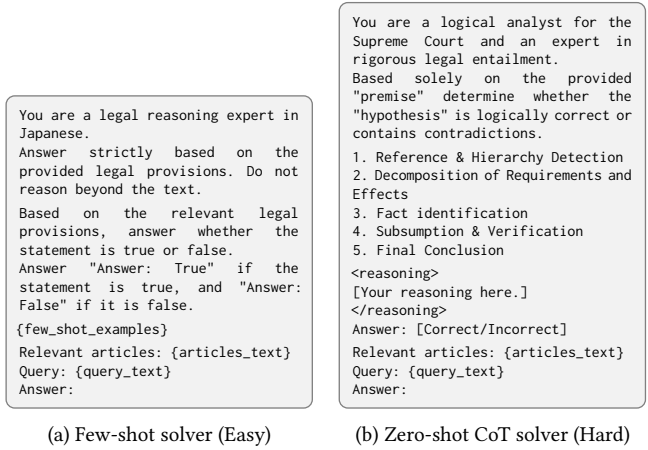

    \centering
    \begin{minipage}[t]{0.48\columnwidth}
    \begin{tcolorbox}[
      colback=gray!10, colframe=black!50, boxrule=0.5pt, arc=3pt,
      left=2pt, right=2pt, top=2pt, bottom=2pt, fontupper=\scriptsize\ttfamily
    ]
    You are a legal reasoning expert in Japanese.\\
    Answer strictly based on the provided legal provisions. Do not reason beyond the text.\\[0.3em]
    Based on the relevant legal provisions, answer whether the statement is true or false.\\
    Answer "Answer: True" if the statement is true, and "Answer: False" if it is false.\\[0.3em]
    \{few\_shot\_examples\}\\[0.3em]
    Relevant articles: \{articles\_text\}\\
    Query: \{query\_text\}\\
    Answer:
    \end{tcolorbox}
    \centering\small (a) Few-shot solver (Easy)
    \end{minipage}\hfill
    \begin{minipage}[t]{0.48\columnwidth}
    \begin{tcolorbox}[
      colback=gray!10, colframe=black!50, boxrule=0.5pt, arc=3pt,
      left=2pt, right=2pt, top=2pt, bottom=2pt, fontupper=\scriptsize\ttfamily
    ]
    You are a logical analyst for the Supreme Court and an expert in rigorous legal entailment.\\
    Based solely on the provided "premise" determine whether the "hypothesis" is logically correct or contains contradictions.\\[0.3em]
    1. Reference \& Hierarchy Detection\\
    2. Decomposition of Requirements and Effects\\
    3. Fact identification\\
    4. Subsumption \& Verification\\
    5. Final Conclusion\\[0.3em]
    <reasoning>\\
    {[}Your reasoning here.{]}\\
    </reasoning>\\
    Answer: [Correct/Incorrect]\\[0.3em]
    Relevant articles: \{articles\_text\}\\
    Query: \{query\_text\}\\
    Answer:
    \end{tcolorbox}
    \centering\small (b) Zero-shot CoT solver (Hard)
    \end{minipage}
    \caption{Prompt templates for Task~4 solvers: (a)~balanced few-shot for ``Easy'' cases, (b)~structured chain-of-thought for ``Hard'' cases.}
    \Description{Prompt templates for Task 4 solvers showing few-shot and chain-of-thought approaches.}
    \label{fig:task4-solvers}
    \end{figure}
\end{itemize}

\subsection{Experimental Setup}
To evaluate the individual and combined contributions of our proposed modules, we structured our submission into three distinct runs. Figure~\ref{fig:task4-solvers} shows the prompt templates used by both solvers. This setup allows us to compare specialized single-strategy approaches against our dynamic routing methodology:

\begin{itemize}
    \item \textbf{Run 1 (Balanced Few-Shot):} A simplified approach bypassing the difficulty classifier entirely. All queries, regardless of their inherent complexity, are processed directly by the Dynamic Balanced Few-Shot Solver ($k=6$). All inference in this run is powered by \textit{DeepSeek-V3\footnote{\url{https://huggingface.co/deepseek-ai/DeepSeek-V3-0324}}}. This serves as our foundational baseline for in-context pattern matching.

    \item \textbf{Run 2 (Pure Zero-Shot CoT):} This run also bypasses the routing mechanism, instead applying the highly structured Zero-Shot Chain-of-Thought Solver to every query in the test set. Powered entirely by \textit{DeepSeek-V3\footnotemark[11]}, this configuration evaluates the model's raw logical deduction and rule extraction capabilities without the influence or aid of retrieved few-shot examples.

    \item \textbf{Run 3 (Hybrid Routing Pipeline):} This run deploys our complete proposed methodology. To optimize for both operational efficiency and reasoning depth, we employ a mixed-model architecture. The Difficulty Classifier utilizes \textit{Llama-3.3-70b-versatile} for rapid categorization. Based on this assessment, queries are dynamically routed to either the Few-Shot Solver (for ``Easy'' cases) or the CoT Solver (for ``Hard'' cases). \textit{DeepSeek-V3\footnotemark[11]} serves as the core reasoning engine for both solvers. This run tests the overall efficacy of dynamically adapting the prompting strategy based on query complexity.
\end{itemize}

 \subsection{Results and Discussion}

The official evaluation results for COLIEE Task 4 are summarized in Table \ref{tab:results}. In this highly competitive task, our team (\textbf{NOWJ}) achieved a Top 7 standing among all participating institutions. 

Table \ref{tab:results} presents a streamlined version of the official leaderboard, displaying one representative run per rank alongside all three of our submitted configurations (NOWJ-1, NOWJ-2, and NOWJ-3). Our best-performing configuration, \textbf{NOWJ-1}, secured Rank 8 overall with an accuracy of 0.8780 (72 correct predictions).

\begin{table}[h]
\centering
\caption{Official Leaderboard for COLIEE Task 4 (Showing top-performing teams and our submissions).}
\label{tab:results}
\begin{tabular}{lccc}
\toprule
\textbf{Team} & \textbf{Run} & \textbf{Accuracy} & \textbf{Correct} \\
\midrule
DU & DU1 & \textbf{0.9634} & \textbf{79} \\
JNLP & JNLP-3 & 0.9512 & 78 \\
IAI & IAIrun1 & 0.9390 & 77 \\
RUG & RUG-1 & 0.9268 & 76 \\
UA & UA1 & 0.9146 & 75 \\
cody & cody5 & 0.9024 & 74 \\
{\ul NOWJ} & {\ul NOWJ-1} & {\ul 0.8780} & {\ul 72} \\
HUKB & HUKBac & 0.8780 & 72 \\
{\ul NOWJ} & {\ul NOWJ-3} & {\ul 0.8049} & {\ul 66} \\
AiR & AiIRCrossCoT & 0.7805 & 64 \\
Bengo & BengoREF & 0.7317 & 60 \\
ASHIN & ASHIN-LLAMA3 & 0.6098 & 50 \\
{\ul NOWJ} & {\ul NOWJ-2} & {\ul 0.5000} & {\ul 41} \\
\bottomrule
\end{tabular}
\end{table}

To validate our methodology prior to the final submission, we conducted retrospective evaluations on historical COLIEE test sets from 2018, 2019, and 2020. The performance of our three configurations on these datasets is presented in Table \ref{tab:historical_results}.

\begin{table}[h]
\centering
\caption{Accuracy of the proposed configurations on historical COLIEE test sets.}
\label{tab:historical_results}
\begin{tabular}{lccc}
\toprule
\textbf{Dataset} & \textbf{Run 1} & \textbf{Run 2} & \textbf{Run 3} \\
\midrule
COLIEE 2018 & 0.8060 & 0.7761 & \textbf{0.9104} \\
COLIEE 2019 & 0.7909 & 0.7818 & \textbf{0.8273} \\
COLIEE 2020 & 0.8889 & 0.9012 & \textbf{0.9136} \\
\bottomrule
\end{tabular}
\end{table}

As shown in Table \ref{tab:historical_results}, there is a clear performance hierarchy on the historical datasets: Run 3 consistently outperforms both Run 1 and Run 2. This strongly supported our initial hypothesis that a dynamic routing pipeline leveraging both few-shot pattern matching and rigorous CoT deduction is superior to any single-strategy approach. Furthermore, the Pure CoT approach (Run 2) generally underperformed the Few-Shot baseline (Run 1), indicating that strict logical deduction without examples is inherently challenging.

While the historical data favored the Hybrid Pipeline (Run 3), the current year’s blind test reveals a shift: the Few-Shot baseline (Run 1) is now the top performer ($87.80\%$), significantly outperforming both the Hybrid model ($80.49\%$) and the underperforming Pure CoT configuration (Run 2) ($50.00\%$). The performance drop in Run 2 and 3 suggests two primary issues:
\begin{itemize}
    \item \textbf{Rigidity of the CoT Guardrails:} The strict logical guardrails in the Zero-Shot CoT prompt, while preventing hallucinations on older data, likely failed to account for new semantic nuances, causing the model to default to "Incorrect/Neutral.".
    \item \textbf{Robustness of In-Context Learning over Strict Deduction:} Few-Shot in-context learning proved more resilient than isolated deduction. The retrieved examples provided a more reliable signal for "Hard" queries than the rigid step-by-step logic imposed by the CoT prompt.
\end{itemize}

Consequently, future work will shift from zero-shot prompting toward domain-specific fine-tuning. By training an LLM explicitly on complex legal cases, we aim to replace prompt-based guardrails with fine-tuned reasoning capabilities.

\section{Pilot Task: Legal Judgment Prediction for Japanese Tort cases}
\subsection{Task Overview}
This pilot task focuses on legal judgment prediction for Japanese civil cases concerning torts. In these scenarios, plaintiffs argue that a defendant's action constitutes a tort, while defendants contest the plaintiffs' arguments. The overarching objective is to process a set of case inputs to generate specific judicial outputs across two interconnected tasks: Tort Prediction and Rationale Extraction. Formally, the systems take $(U, P, D)$ as input, where $U$ represents undisputed facts that are agreed upon or not disputed by any parties, $P$ denotes the arguments from the plaintiffs, and $D$ encompasses the arguments from the defendants. Given these inputs, systems must output $(T, R^P, R^D)$, defined by the following tasks:
\begin{itemize}
\item \textbf{Tort Prediction (TP)}: Predicts whether a tort is affirmed ($T$). Here, $T$ is a Boolean value representing the final decision. Model performance for this task is evaluated using \textit{accuracy}.
\item \textbf{Rationale Extraction (RE)}: The final decision ($T$) is based on the arguments accepted by the judge, making these accepted arguments the rationales for the decision. RE identifies these accepted arguments by outputting $R^P$ (for plaintiffs) and $R^D$ (for defendants). Both are sequences of Boolean values denoting accepted arguments as True within $P$ and $D$. This task is evaluated using the \textit{micro-F1} measure.
\end{itemize}

\subsection{Methodology}
Table~\ref{tab:pilot_task_analysis} presents key dataset statistics. The training and test sets show consistent distributions, with average claims per party stable around 3.5–3.9 claims. However, maximum complexity varies substantially: some cases contain over 100 plaintiff claims or more than 140 undisputed facts, requiring the system to handle both simple and highly complex cases. The balanced label distribution in training data, approximately 51\% acceptance for both parties, indicates no inherent bias toward accepting or rejecting claims.

\begin{table}[ht]
\caption{Dataset characteristics for the Legal Judgment Prediction task across training and test sets.}
\label{tab:pilot_task_analysis}
\centering
\begin{tabular}{@{}lccc@{}}
\toprule
\textbf{Characteristic}        & \textbf{Train} & \textbf{Test 2025} & \textbf{Test 2026} \\ \midrule
Cases                          & 6,508    & 812      & 803      \\
Avg. facts                     & 1.33     & 1.37     & 1.30     \\
Max. facts                     & 134      & 26       & 141      \\
Avg. plaintiff claims          & 3.88     & 3.83     & 3.77     \\
Max. plaintiff claims          & 111      & 130      & 125      \\
Avg. defendant claims          & 3.45     & 3.50     & 3.78     \\
Max. defendant claims          & 86       & 50       & 99       \\ \bottomrule
\end{tabular}
\end{table}

\begin{figure*}[h!]
    \centering
    \includegraphics[width=1\linewidth]{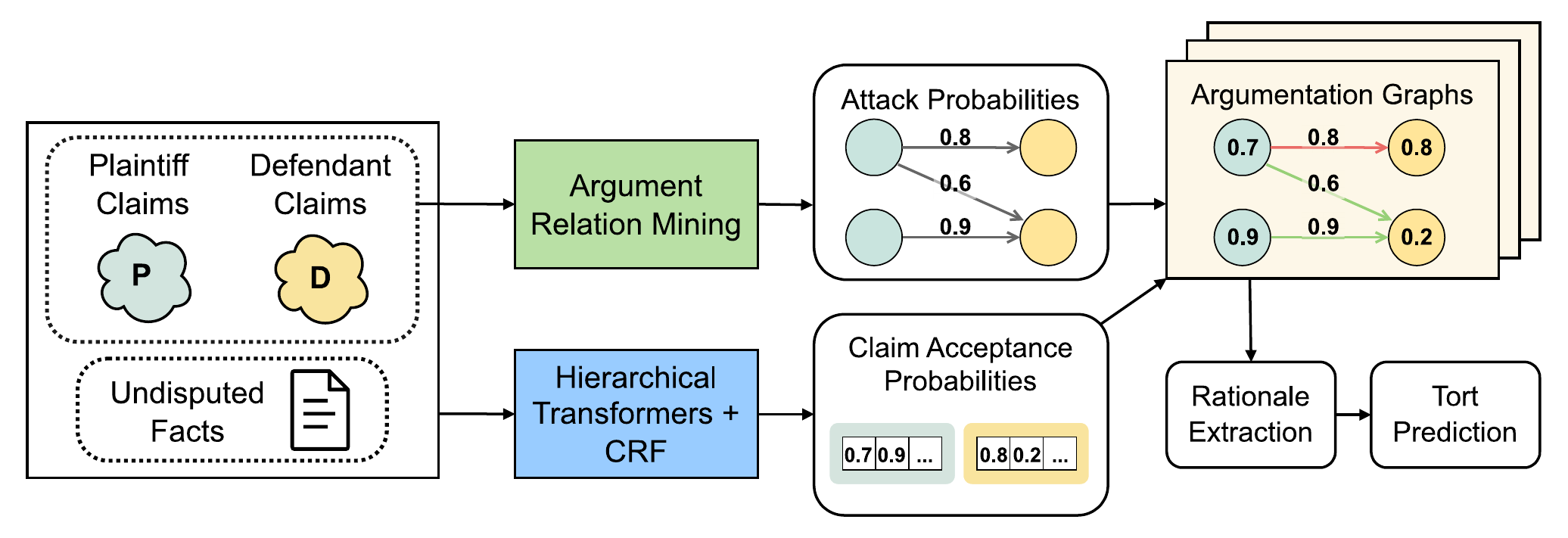}
    \caption{System architecture combining neural claim assessment and argumentation for legal judgment prediction.}
    \label{fig:task5_framework}
    \Description{Task 5 framework}
\end{figure*}

Our system, illustrated in Figure~\ref{fig:task5_framework}, combines neural claim assessment with argumentation-based reasoning. A key feature of this framework is its explainability: the final tort decision follows from the set of accepted claims identified through structured reasoning over attack relations between arguments. Moreover, the system achieves strong performance without relying on large language models, making it computationally efficient and fast to deploy.

\subsubsection{Hierarchical Transformer}
The hierarchical transformer uses a two-stage architecture for claim assessment. The first stage employs a pre-trained Japanese language model as the word-level encoder, processing each claim and the fact sequence independently to generate contextualized embeddings. The second stage uses a span-level transformer that processes claim-level representations combining four information sources: the claim embedding from the word encoder, a positional embedding of the claim's sequential position, a party-type embedding indicating plaintiff or defendant, and the representation of undisputed facts shared across all claims.

The model uses task-specific output layers. For Tort Prediction, a linear layer produces a binary court decision. For Rationale Extraction, we add a Conditional Random Field (CRF) layer above claim representations to model dependencies between claims from the same party, ensuring consistent label sequences. The training loss combines both tasks with a weighted sum to balance the two objectives. The model outputs binary predictions for the court decision via the TP head, binary label sequences for accepted claims via CRF decoding, and continuous probability scores for each claim's acceptance derived from \textit{softmax} on the rationale logits. These probability scores serve as input to the argumentation reasoning module for conflict resolution.

\subsubsection{Argument Relation Mining}
To identify conflicts between opposing claims, we employ an argument relation mining model. We train the model on Japanese-translated argument mining datasets. The model analyzes all plaintiff-defendant claim pairs and performs binary classification to determine whether claims conflict with each other. We extract conflict relations above a confidence threshold as attacks, where one claim contradicts another. This produces a directed attack graph with confidence scores that represent the likelihood of each conflict relation.

\subsubsection{Argumentation-Based Reasoning}
This module resolves conflicts through structured reasoning over multiple graph configurations. The reasoning process operates in three stages. First, \textit{graph ensemble generation} addresses uncertainty in attack detection. For $N$ detected attacks, up to $2^N$ possible graphs exist depending on which attacks are valid. Rather than enumerating all possibilities, we use a priority queue to construct the top-$K$ most probable graphs, where each graph $\mathcal{G}_\ell$ is assigned probability $P(\mathcal{G}_\ell)$ based on the likelihood of its included and excluded attacks. Second, \textit{attack resolution} determines claim acceptance within each graph. An attack from claim $a$ to claim $b$ succeeds if $a$ has equal or higher acceptance probability than $b$ from the hierarchical transformer. Claims defeated by successful attacks are rejected, while unattacked claims or those surviving failed attacks retain their predicted status. Third, \textit{probabilistic aggregation} combines results across the $K$ graphs through weighted voting. For each claim $c$, we compute:
\[
\rho(c) = \sum_{\ell=1}^K w_\ell \cdot v_\ell(c),
\]
where $v_\ell(c) \in \{0,1\}$ indicates whether $c$ is accepted in graph $\mathcal{G}_\ell$, and $w_\ell$ are normalized graph probabilities satisfying $\sum_{\ell=1}^K w_\ell = 1$. The value $\rho(c)$ represents the total probability weight of graphs accepting $c$. A claim is accepted if $\rho(c) \geq 0.5$.

After determining accepted claims for both parties, the tort decision is made through a counting mechanism: the plaintiff prevails if they have more accepted claims than the defendant, reflecting the adversarial nature of tort litigation. This reasoning applies only to cases with detected conflicts; cases without attacks use direct predictions from the hierarchical transformer. The entire framework operates without LLMs, relying instead on a compact hierarchical transformer and lightweight reasoning components that enable rapid inference.

\subsection{Experimental Setup}
We implement the hierarchical transformer using \textit{ModernBERT-Ja-310M}\footnote{\url{https://huggingface.co/sbintuitions/modernbert-ja-310m}} as the word-level encoder. Sequence limits are set to 64 claims per case, 64 tokens per claim, and 512 tokens for concatenated facts. Missing components are handled by substituting empty sequences. The model is trained with a multi-task loss: $\mathcal{L}_{\text{total}} = 0.6 \cdot \mathcal{L}(\text{TP}) + 0.4 \cdot \mathcal{L}(\text{RE})$, where greater weight is assigned to tort prediction.

For argument relation mining, we finetune a pre-trained model\footnote{\url{https://huggingface.co/raruidol/ArgumentMining-EN-ARI-US2016}} on Japanese-translated argument mining datasets, extracting conflict relations with a confidence threshold of 0.5. For argumentation reasoning, we use $K=5$ graphs. Post-processing is applied to rationale predictions based on party-level consistency: if accepted claims outnumber rejected claims by a factor of $x=1.5$, all claims from that party are set to accepted; if rejected claims outnumber accepted claims by the same factor, all claims are set to rejected. Cases without a clear majority remain unchanged.

We submit three configurations to evaluate component contributions: \textit{Run 1} uses argumentation reasoning for TP and hierarchical transformer predictions for RE; \textit{Run 2} adds post-processing to RE; \textit{Run 3} uses only the hierarchical transformer as an ablation baseline.

\subsection{Results and Discussion}
Table~\ref{tab:pilot_results} shows official results on the COLIEE 2026 test set. Our system achieved 3rd place for both tasks. For TP, our best runs (Runs 1–2) achieved 67.6\% accuracy, while for RE, our best run (Run 2) achieved 64.8\% F1.

\begin{table}[h!]
\caption{Official results for the COLIEE 2026 Pilot Task.}
\centering
\label{tab:pilot_results}
\begin{subtable}[t]{0.22\textwidth}
\centering
\begin{tabular}{lc}
\hline
\textbf{Team}      & \textbf{Accuracy} \\ \hline
JNLP               & \textbf{72.7\%}   \\
UIT                & 72.6\%            \\
{\ul NOWJ (run 1)} & {\ul 67.6\%}      \\
{\ul NOWJ (run 2)} & {\ul 67.6\%}      \\
{\ul NOWJ (run 3)} & {\ul 67.1\%}      \\
KIS                & 66.6\%            \\ \hline
\end{tabular}
\subcaption{Tort Prediction}
\label{tab:tp}
\end{subtable}%
\hfill
\begin{subtable}[t]{0.22\textwidth}
\centering
\begin{tabular}{lc}
\hline
\textbf{Team}      & \textbf{F1 score} \\ \hline
UIT                & \textbf{68.2\%}   \\
JNLP               & 67.4\%            \\
{\ul NOWJ (run 2)} & {\ul 64.8\%}      \\
{\ul NOWJ (run 3)} & {\ul 63.4\%}      \\
{\ul NOWJ (run 1)} & {\ul 63.4\%}      \\
KIS                & 62.6\%            \\ \hline
\end{tabular}
\subcaption{Rationale Extraction}
\label{tab:re}
\end{subtable}
\end{table}

Comparing the three runs reveals each component's contribution. For TP, Runs 1 and 2, which incorporate argumentation reasoning, achieve 67.6\% accuracy, outperforming Run 3 (67.1\%) by 0.5 percentage points. For RE, Run 2, which adds post-processing, achieves 64.8\% F1, outperforming Runs 1 and 3 (both 63.4\%) by 1.4 percentage points. The performance gap to the top team is 5.1 percentage points for TP and 3.4 percentage points for RE, presenting clear opportunities for future improvement. The argumentation reasoning component contributed a 0.5 percentage point gain in TP accuracy, suggesting that conflict modeling is a promising direction for further development.

A key strength of our framework lies in its explainability. Unlike black-box approaches, our system produces decisions that are traceable: the tort outcome follows directly from the set of accepted claims identified through explicit argumentation graphs. This transparency enables practitioners to inspect which claims were accepted or rejected and how conflicts between opposing arguments were resolved. Furthermore, the framework achieves competitive performance without relying on large language models, instead using a compact hierarchical transformer and lightweight reasoning components that enable rapid inference with modest computational requirements. Future work to enhance performance includes several directions. First, training argument relation models on native Japanese data to better capture language-specific argumentation patterns. Second, incorporating richer attack resolution strategies that consider claim semantics and supporting evidence, and integrating undisputed facts directly into the reasoning stage to ground conflict resolution in factual context. Third, exploring alternative argumentation semantics such as preferred or stable extensions, and integrating external legal knowledge to further improve prediction accuracy while maintaining the interpretability that our framework provides.

\section{Conclusion}
This paper described the NOWJ team's participation in all five tasks of COLIEE 2026, achieving 1st place on Task~1 (F1: 0.4220) and Task~3 (accuracy: 0.9512), 4th on Task~2 (F1: 0.4429), 8th on Task~4 (accuracy: 0.8780), and 3rd on the Pilot Task (Accuracy: 67.6\%, F1: 64.8\%). Key findings include: MLP-based reranking with handcrafted features outperformed generative cross-encoders (Task~1), LLM consensus verification yields high precision but limited recall (Task~2), attention-based reranking via QRHead substantially improves LLM reasoning accuracy (Task~3), difficulty-based routing is vulnerable to distribution shift (Task~4), and argumentation graph reasoning provides modest but explainable gains for tort prediction (Pilot Task). Future work will focus on improving retrieval recall through domain-specific fine-tuning, relaxing entailment verification thresholds, decoupling article filtering from entailment prediction, replacing zero-shot prompting with fine-tuned models, and training argument relation classifiers on Japanese data.





\bibliographystyle{ACM-Reference-Format}
\bibliography{sample-base}

@String{Computing = "Computing" }

@String{Springer = "Springer-Verlag" }

@inproceedings{khattab2020colbert,
author = {Khattab, Omar and Zaharia, Matei},
title = {ColBERT: Efficient and Effective Passage Search via Contextualized Late Interaction over BERT},
year = {2020},
isbn = {9781450380164},
publisher = {Association for Computing Machinery},
address = {New York, NY, USA},
url = {https://doi.org/10.1145/3397271.3401075},
doi = {10.1145/3397271.3401075},
booktitle = {Proceedings of the 43rd International ACM SIGIR Conference on Research and Development in Information Retrieval},
pages = {39–48},
numpages = {10},
location = {Virtual Event, China},
series = {SIGIR '20}
}

@misc{msswift,
      title={SWIFT:A Scalable lightWeight Infrastructure for Fine-Tuning}, 
      author={Yuze Zhao and Jintao Huang and Jinghan Hu and Xingjun Wang and Yunlin Mao and Daoze Zhang and Hong Zhang and Zeyinzi Jiang and Zhikai Wu and Baole Ai and Ang Wang and Wenmeng Zhou and Yingda Chen},
      year={2025},
      eprint={2408.05517},
      archivePrefix={arXiv},
      primaryClass={cs.CL},
      url={https://arxiv.org/abs/2408.05517}, 
}

@misc{qwen3reranker,
      title={Qwen3 Embedding: Advancing Text Embedding and Reranking Through Foundation Models}, 
      author={Yanzhao Zhang and Mingxin Li and Dingkun Long and Xin Zhang and Huan Lin and Baosong Yang and Pengjun Xie and An Yang and Dayiheng Liu and Junyang Lin and Fei Huang and Jingren Zhou},
      year={2025},
      eprint={2506.05176},
      archivePrefix={arXiv},
      primaryClass={cs.CL},
      url={https://arxiv.org/abs/2506.05176}, 
}

@misc{hu2021lora,
      title={LoRA: Low-Rank Adaptation of Large Language Models}, 
      author={Edward J. Hu and Yelong Shen and Phillip Wallis and Zeyuan Allen-Zhu and Yuanzhi Li and Shean Wang and Lu Wang and Weizhu Chen},
      year={2021},
      eprint={2106.09685},
      archivePrefix={arXiv},
      primaryClass={cs.CL},
      url={https://arxiv.org/abs/2106.09685}, 
}

@misc{uqlegalai2025,
      title={UQLegalAI@COLIEE2025: Advancing Legal Case Retrieval with Large Language Models and Graph Neural Networks}, 
      author={Yanran Tang and Ruihong Qiu and Zi Huang},
      year={2025},
      eprint={2505.20743},
      archivePrefix={arXiv},
      primaryClass={cs.IR},
      url={https://arxiv.org/abs/2505.20743}, 
}

@inproceedings{jnlp2025,
  title={JNLP@COLIEE 2025: Hybrid Large Language Model-based Framework for Legal Information Retrieval and Entailment},
  author={Nguyen, Hai and Nguyen, Hiep and Pham, Trang and Nguyen, Minh and Trieu, An and Do, Dinh-Truong and Le, Nguyen-Khang and Nguyen, Le-Minh},
  booktitle={Proceedings of the Workshop on the Twelfth International Competition on Legal Information Extraction and Entailment (COLIEE 2025)},
  year={2025},
  address={Chicago, USA},
  publisher={ACM},
  pages={57--66}
}

@inproceedings{umnlp2024,
author = {Curran, Damian and Conway, Mike},
title = {Similarity Ranking of Case Law Using Propositions as Features},
year = {2024},
isbn = {978-981-97-3075-9},
publisher = {Springer-Verlag},
address = {Berlin, Heidelberg},
url = {https://doi.org/10.1007/978-981-97-3076-6_11},
doi = {10.1007/978-981-97-3076-6_11},
booktitle = {New Frontiers in Artificial Intelligence: JSAI International Symposium on Artificial Intelligence, JSAI-IsAI 2024, Hamamatsu, Japan, May 28–29, 2024, Proceedings},
pages = {156–166},
numpages = {11},
location = {Hamamatsu, Japan}
}

@inproceedings{zhang2025query,
    title = "Query-Focused Retrieval Heads Improve Long-Context Reasoning and Re-ranking",
    author = "Zhang, Wuwei  and
      Yin, Fangcong  and
      Yen, Howard  and
      Chen, Danqi  and
      Ye, Xi",
    editor = "Christodoulopoulos, Christos  and
      Chakraborty, Tanmoy  and
      Rose, Carolyn  and
      Peng, Violet",
    booktitle = "Proceedings of the 2025 Conference on Empirical Methods in Natural Language Processing",
    month = nov,
    year = "2025",
    address = "Suzhou, China",
    publisher = "Association for Computational Linguistics",
    url = "https://aclanthology.org/2025.emnlp-main.1214/",
    doi = "10.18653/v1/2025.emnlp-main.1214",
    pages = "23791--23805",
    ISBN = "979-8-89176-332-6"
}

@inproceedings{goebel2025overview,
author = {Goebel, Randy and Kano, Yoshinobu and Kim, Mi-Young and Kwan, Calum and Satoh, Ken and Yamada, Hiroaki and Yoshioka, Masaharu},
title = {An Overview of the COLIEE 2025 Competition: Legal Case Law and Statute Law Information Retrieval and Entailment},
year = {2026},
isbn = {9798400719394},
publisher = {Association for Computing Machinery},
address = {New York, NY, USA},
url = {https://doi.org/10.1145/3769126.3785016},
doi = {10.1145/3769126.3785016},
booktitle = {Proceedings of the Twentieth International Conference on Artificial Intelligence and Law},
pages = {506–515},
numpages = {10},
location = {
},
series = {ICAIL '25}
}

@article{vuong2025uncovering,
  title={Uncovering connections: a reference network approach to statute law retrieval},
  author={Vuong, Thi-Hai-Yen and Nguyen, Hai-Long and Nguyen, Tan-Minh and Nguyen, Ha-Thanh and Nguyen, Le-Minh and Phan, Xuan-Hieu},
  journal={Applied Intelligence},
  volume={55},
  number={13},
  pages={936},
  year={2025},
  publisher={Springer}
}

@article{kis2025,
  author    = {Onaga, Takaaki and Kano, Yoshinobu},
  title     = {KIS: COLIEE 2025 Task 4 Solver Using Japanese LLM},
  journal   = {The Review of Socionetwork Strategies},
  year      = {2026},
  date      = {2026/03/18},
  volume    = {20},
  pages     = {341--359},
  doi       = {10.1007/s12626-026-00209-w},
  url       = {https://doi.org/10.1007/s12626-026-00209-w},
  issn      = {1867-3236}
}

@article{captain2025,
    author = {Nguyen, Dat and Nguyen, Minh-Phuong and Chu, Quang-Huy and Luu, Son T. and Chu, Nguyen-Hoang and Vo, Trung and Nguyen, Le-Minh}, 
    title = {Enhancing Legal Text Processing and Structural Analysis with Large Language Models at {COLIEE} 2025}, 
    journal = {The Review of Socionetwork Strategies}, 
    year = {2026}, 
    volume = {20}, 
    number = {1}, 
    pages = {361--383}, 
    issn = {1867-3236}, 
    doi = {10.1007/s12626-026-00211-2}, 
    url = {https://doi.org/10.1007/s12626-026-00211-2} 
}

@article{rabelo2024coliee,
  author    = {Juliano Rabelo and Randy Goebel and Mi-Young Kim and Yoshinobu Kano and Masaharu Yoshioka and Ken Satoh},
  title     = {Overview and Discussion of the Competition on Legal Information Extraction/Entailment ({COLIEE}) 2023},
  journal   = {The Review of Socionetwork Strategies},
  volume    = {18},
  number    = {1},
  pages     = {27--47},
  year      = {2024}
}

\appendix









\end{document}